\documentclass[conference]{IEEEtran}
\IEEEoverridecommandlockouts

\usepackage{cite}
\usepackage{amsmath,amssymb,amsfonts}
\usepackage{algorithmic}
\usepackage{efbox,graphicx}
\usepackage{textcomp}
\usepackage{xcolor}
\usepackage{multirow}

\graphicspath{ {./Fused/p112/} }
\graphicspath{ {results/} }

\def\BibTeX{{\rm B\kern-.05em{\sc i\kern-.025em b}\kern-.08em
    T\kern-.1667em\lower.7ex\hbox{E}\kern-.125emX}}
    
\efboxsetup{linecolor=blue,linewidth=0.5pt}    
    
\newcommand{\rectangle}{\fboxsep0pt\fbox{\rule{0.5em}{0pt}\rule{0pt}{1.6ex}}}

\begin{document}

\title{Multi-channel MRI Embedding: An Effective Strategy for Enhancement of Human Brain Whole Tumor Segmentation}

\author{Apurva Pandya, Catherine Samuel, Nisargkumar Patel, Vaibhavkumar Patel, and Thangarajah~Akilan,~\IEEEmembership{Member,~IEEE}%

% \thanks{A. Pandya, C. Samuel, N. Patel, V. Patel, and T. Akilan are with the Department of Computer Science, Lakehead University, Thunder Bay, ON, Canada. (e-mail: \{apandya1, csamuel1, npatel34, vpatel26, takilan\}@lakeheadu.ca).}

}

\maketitle

\begin{abstract}
One of the most important tasks in medical image processing is the brain's whole tumor segmentation. It assists in quicker clinical assessment and early detection of brain tumors, which is crucial for lifesaving treatment procedures of patients. Because, brain tumors often can be malignant or benign, if they are detected at an early stage. A brain tumor is a collection or a mass of abnormal cells in the brain. The human skull encloses the brain very rigidly and any growth inside this restricted place can cause severe health issues. The detection of brain tumors requires careful and intricate analysis for surgical planning and treatment. Most physicians employ Magnetic Resonance Imaging (MRI) to diagnose such tumors. 
A manual diagnosis of the tumors using MRI is known to be time-consuming; approximately, it takes up to eighteen hours per sample. Thus, the automatic segmentation of tumors has become an optimal solution for this problem.
Studies have shown that this technique provides better accuracy and it is faster than manual analysis resulting in patients receiving the treatment at the right time. Our research introduces an efficient strategy called Multi-channel MRI embedding to improve the result of deep learning-based tumor segmentation. The experimental analysis on the Brats-2019 dataset wrt the U-Net encoder-decoder (EnDec) model shows significant improvement. The embedding strategy surmounts the state-of-the-art approaches with an improvement of 2\% without any timing overheads. 
\end{abstract}


%-----------------------
\section{Introduction}
%-----------------------

Segmentation of brain tumors using MRI remains a challenging task, even with human intervention. Because, brain tumors can have vastly different sizes, shape and morphology, and can appear anywhere in the brain, the images are often poorly diffused and contrasted in nature. Brain tumors can be very lethal and early detection of these cancers are necessary to improve the treatments of patients. Amongst the different types, Gliomas are one of the most common brain tumors. Gliomas are mostly graded into low-grade and high-grade Gliomas namely LGG and HGG~\cite{claus2015survival,schapira2003communication}. The HGG are the most aggressive and dangerous ones, as they are most highly invasive tumors that grow aggressively and instantly invade the Central Nervous System.  Therefore, such Gliomas need to be detected early to increase the life expectancy of the patients~\cite{schapira2003communication}. According to a survey conducted by the US National Cancer Institute (NCI), there are approximately $18,000$ Americans diagnosed with a Glioma brain tumor yearly and most of them expired within 14 months~\cite{schapira2003communication,leece2017global}. With drastic advancement in the recent clinical practices, medical imaging, MRI, and Computed Tomography (CT) have been extensively used to determine: (i). the presence of a tumor, (ii). the spread into other locations such as Central Nervous System, and (iii). the detection of edemas. 
Here, the magnetic resonance imaging is a non-invasive technology that produces three dimensional detailed anatomical images of the human body. It has been a gold-standard imaging modality for diagnosis and treatment planning/monitoring of the brain because of its superior image contrast in soft tissues and higher sensitivity~\cite{huo2015label}. 

\begin{figure}[!tbp]
  \centering
    \includegraphics[width=0.49\textwidth]{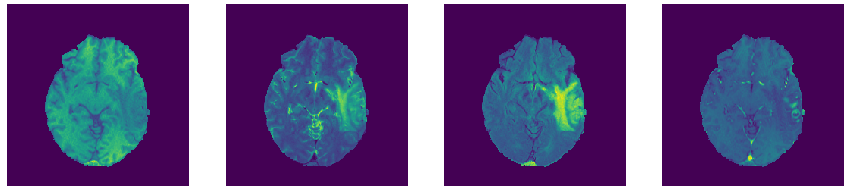}\\
    \vspace{-0.05cm}
   \small{T1\hspace{1.9cm} T2\hspace{1.5cm} FLAIR\hspace{1.3cm} T1CE\hspace{-0.5cm} }
  \caption{Types of MRI modalities: T1-weighted, T2-weighted, T1ce and FLAIR modalities of $125^{th}$ brain slice of a single patient from BraTS'19 dataset~\cite{menze2014multimodal}.}
  
  \label{fig:mri_modalities}
  \vspace{-0.5cm}
\end{figure}

Although MRI provides better quality data than regular X-Rays or CT scans, the complexity of the brain tumors and the rapid multiplicity of the cancer cells often makes the tumor recognition and segmentation task quite difficult for radiologists and medical clinicians. 
The four majorly known standard MRI modalities are T1-weighted MRI (T1), T2-weighted MRI (T2), T1-weighted MRI with gadolinium contrast enhancement (T1-Gd), and Fluid Attenuated Inversion Recovery (FLAIR)~\cite{liu2019recent}. Fig.~\ref{fig:mri_modalities} depicts samples of the different MRI modalities collected from Brats-2019 dataset.
Generally, distinguishing the healthy tissues from the cancer prone area is done using T1, whereas T2 is used to distinguish the edema region with a bright signal intensity. In T1-Gd modality, the tumor border can easily be identified using the accumulated contrast agent (Gadolinium ions) in the active cell region. Similarly, the FLAIR images are mostly used to distinguish the edema region from the Cerebrospinal Fluid (CSF), since the signal of water molecules in the brain are suppressed~\cite{li2018t1}. 

It is found that the manual segmentation is very laborious and it involves a tedious process that leads to inter-expert variability and less accuracy. Consequently, this has brought about an increased usage of automatic segmentation of heterogeneous tumors that highly impacts the clinical medicine field by freeing physicians from
the burden of the manual delineation of the tumors. Typically, with manual diagnosis, the time consumed is approximately eighteen hours, whereas with automatic process, each tumor can be diagnosed within thirty seconds to a minute~\cite{despotovic2015mri}. Furthermore, if computer algorithms, like deep learning (DL) and convolutional neural networks (CNN) can provide accurate measure of tumor detection. Thus, these automated measurements will aid in the increase of diagnosis or treatment of brain tumors~\cite{despotovic2015mri}. 

The automated MRI segmentation is aim to categorize image pixels into semantically meaningful non-overlapping anatomical regions, such as, bones, muscles, and blood vessels. The algorithms are to learn accurate regional divisions based on attributes, like, intensity, depth, color, or texture~\cite{liu2019recent,ronneberger2015u,morey2009comparison}. 

Rest of the organization of this paper as follows. Section~\ref{literature_review} reviews related literature,

%-------------------------------------------
\section{Literature Review}\label{literature_review}
%-------------------------------------------

To study the complex relationships of structures and organs inside a brain, a thorough segmentation of anatomical features on brain MRI is quite crucial. Mainly, the MRI segmentation methods can be grouped into conventional approaches and deep learning-based approaches.  

\subsection{Conventional MRI Segmentation Approaches}

 Automatic analysis and segment labelling of brain MRI can be effectively done by a conventional approaches, such as Gaussian mixture models (GMM)~\cite{chen2016improved}, and multi-atlas segmentation akin to template matching algorithm~\cite{huo2015label}. The multi-atlas method uses multiple example of patches known as ‘Atlases’ that are expert-segmented samples and are typically used for registering with a target image. These examples also have many deformed atlas partitions that are later combined through label fusion. It works based on appearance correlation of image patches. Hence, a target image can be segmented by referring or correlating to the expert-labeled atlases. Some researchers, like Wang~\textit{et al.}~\cite{wang2012multi} introduce a weighted voting mechanism with different weights that are typically derived from the intensity similarity between atlases and target image to improve the segmentation results. The major limitation of those methods is that the weights must be computed individually for each atlas. And so, the researchers propose a way to mitigate the number of similar label errors. This is done by creating a pairwise dependency between the atlases and forming a probability of at least two atlases making a label error at a given voxel (pixels with volume). Thus, the new weighted voting incorporates intensity information by up-weighting atlases that are more similar to the target anatomy in the voting procedure. Since the multi-atlas-based segmentation makes use of more atlases to compute the fused label, it compensates for the potential bias associated with single atlas-based methods. 

 Although the multi-atlas approach achieves more accuracy when compared with single atlas-based segmentation it highly depends on the pairwise atlas-target registration resulting in a huge inconsistency in its spatial distribution and higher computational cost. To mitigate the issue, Huo~\textit{et al.}~\cite{huo2018supervoxel} have proposed a new framework purely based on supervoxels with an estimation of maximum-a-posteriori (MAP) for choosing the label which maximizes the posterior probability. These supervoxels are a collection of voxels that have similar attributes and used to easily replace a voxel grid.

Similarly, in~\cite{huo2015label}, a majority voting scheme that selects the most frequently used labels from the atlases is proposed to avoid the problem of over-segmentation of multi-atlas fusion by comparing the similarity between image patches. 
It ensures that the patch in the atlas with the highest similarity to the target patch is selected. Thus, there is a significant improvement in the anatomical variability by performing separate registrations for each atlas prior to label fusion, which prompts improved robustness against potential pair-wise registration failures.

\subsection{Deep Learning-based Approaches}
%-----------------------------------------

\begin{figure*}[ht]
  \centering
    \includegraphics[trim={0.9cm, 1.9cm, 0.9cm, 1.5cm}, clip, width=\textwidth]{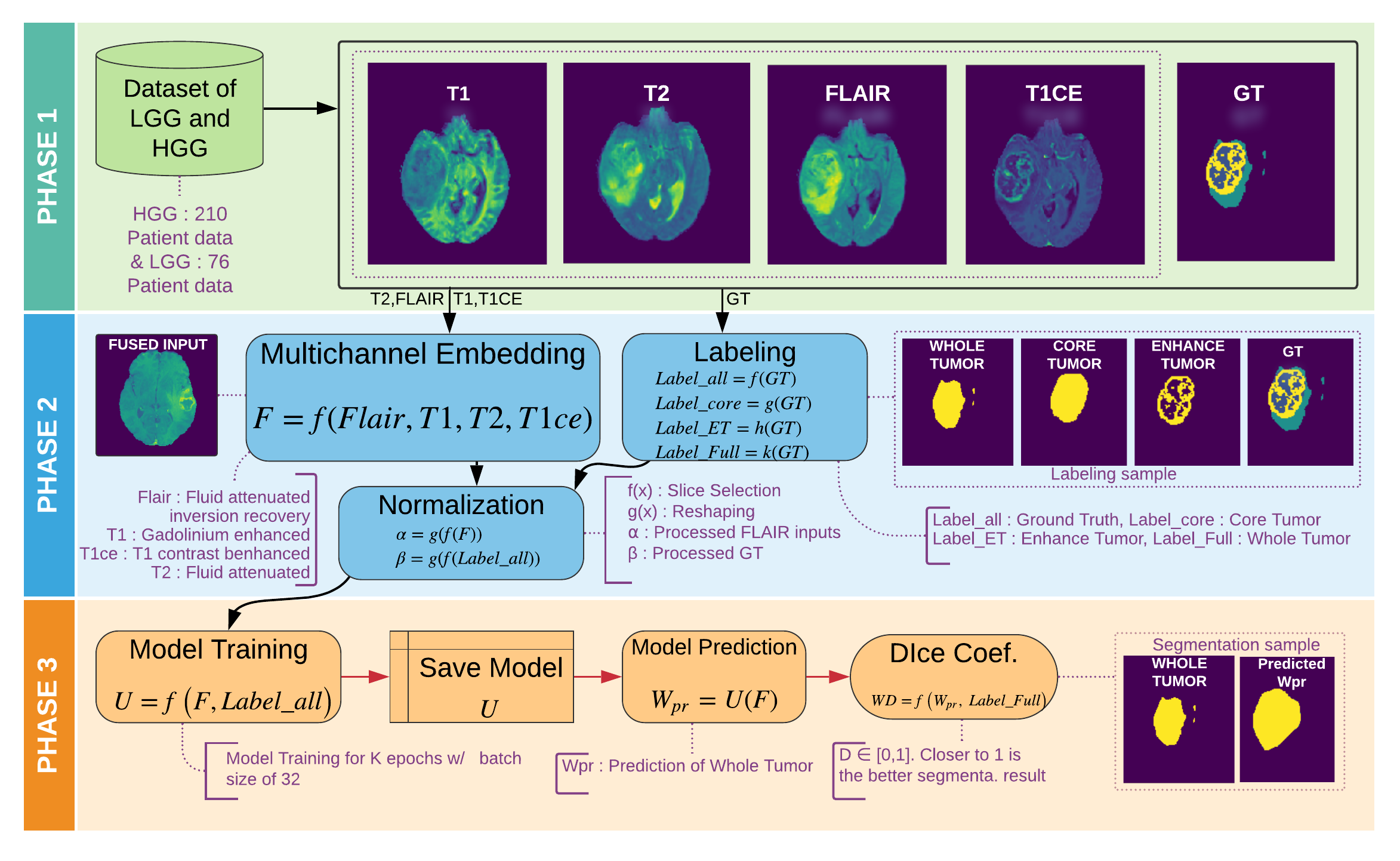}
  \caption{Operational Flow Diagram of the Proposed Multi-channel MRI Embedding-based Brain Whole Tumor Segmentation Model Using Deep Learning.}
  \label{Fig:flow_diagram}
\end{figure*}

Most deep convolutional neural network (CNN) based models for brain tumor segmentation use a 2D/3D patch to predict the class label for the center voxel to identify the tumor~\cite{fang2018three,neppl2019evaluation,feng2018brain}. 
 
The U-Net is a widely used network structure for end-to-end training of MRI segmentation~\cite{8914625, 8913943, 8914497}. It can be used on entire image or extracted patches to provide classification labels over the entire input voxels. Some literature focus on ensemble learning, instead of designing the best single network structure towards improving the segmentation result. For instance, the researchers in \cite{feng2018brain, wang2018two, baid2018deep} propose an ensemble of 3D U-Nets that are trained with different hyper-parameters. 

There are several challenges in directly using the whole images as the input to the 3D U-Net, such as the number of features must be reduced, overfitting, and prolonged training time. Therefore, to utilize the training data more effectively, smaller spatio-volume patches with size of $64 \times 64 \times 4$ are extracted from each subject \cite{feng2018brain}. Generally, the foreground labels contain higher variability. To address this issues, more patches from the foreground voxels are extracted. 

It is shown in \cite{feng2018brain} that the ensemble of all models has the overall best performance as compared with each individual model. 
 
On the other hand, Christian~\textit{et al.}~ \cite{kamnitsas2017efficient} come up with a dual pathway 3D CNN with conditional random field (CRF) to overcome the limitations of standard U-net-based brain lesion segmentation.

The dual pathway architecture incorporates both local and larger contextual information at multiple scales simultaneously. Thus, it effectively removes false positives. Similar to \cite{kamnitsas2017efficient}, the research work conducted by Lin~\textit{et al.}~\cite{ziqi} has incorporated deep medic architecture with a $11$-layer $3D$ CNN that can easily perform multi-scale processing using parallel convolutional paths for the brain tumor segmentation. It uses a Deep Adversarial Network (DAN) comprising of a segmentation network and an evaluation network. Although the training steps of this model is complicated and timing consuming, it records the best result of 0.89 dice-coefficient. 

The major drawback of the 3D U-Net, 3D CNN-based solutions, deep adversarial network, and the CRF refinement stage is, the algorithms are computationally high complex regardless of their slightly enhanced performance over the standard U-Net. Therefore, it is important to explore simple yet effective strategies to improve the basic models. It is the foundation of this work. It introduces an elegant way of improving the segmentation accuracy via embedding of multi-channel MR images, namely T1, T2, FLAIR, and T1ce.

%-----------------------
\section{Proposed Model}
%-----------------------

Figure \ref{Fig:flow_diagram} elaborates the process flow of the proposed multi-channel MRI embedding for enhancement of brain whole tumor segmentation. It subsumes three phases, namely input phase, pre-processing phase, and model training and prediction phase. 

\subsection{Phase 1: Input}
%-------------------------
It takes a dataset consisting of MRI with high grade glioma and low grade gliomas in gzipped NIfTI-1 data format (Neuroimaging Informatics Technology Initiative). For every patient, it has glima information in four different modalities: T1, T1ce, T2, and FLAIR. Since the raw data format cannot be processed by the deep learning model, in this case the U-Net, the NIfTI-1 data must be converted into an appropriate standardized image file format, like Portable Network Graphics (PNG) format without information loss. The converted glima data of all modalities are, then, aligned with the ground truths of the brain whole tumor segmentation. 

 \subsection{Phase 2: Data Preprocessing}
 %-------------------------
 
 \begin{figure*}[!t]
  
    \begin{tabular}{p{0.5cm}p{2cm}p{2cm}p{1.6cm}p{2.0cm}p{1.6cm}p{2cm}p{1.8cm}}
        & \centering $S_1$ & \centering $S_2$ & \centering $S_3$ & \centering $S_4$ & \centering $M_2$ & \centering $M_6$ & \centering $M_9$ \\
    \end{tabular}
    \vspace{-0.5cm}
\begin{center}
%---- SAMPLE 1-----------
     {
    \includegraphics[trim={0.5cm, 0.5cm, 0.2cm, 0.5cm}, clip, width=0.24\columnwidth]{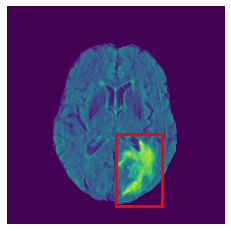}\hspace{0.01cm}
    \includegraphics[trim={0.5cm, 0.5cm, 0.2cm, 0.5cm}, clip,width=0.24\columnwidth]{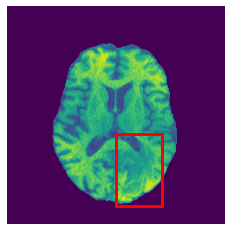}\hspace{0.01cm}
    \includegraphics[trim={0.5cm, 0.5cm, 0.2cm, 0.5cm}, clip,  width=0.24\columnwidth]{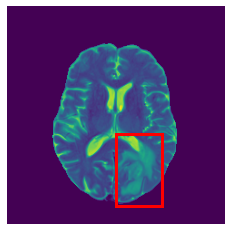}}\hspace{0.01cm}
    {\includegraphics[trim={0.5cm, 0.5cm, 0.2cm, 0.5cm}, clip,  width=0.24\columnwidth]{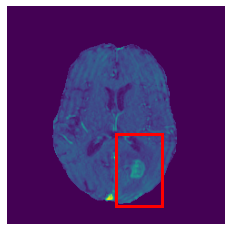}}\hspace{0.01cm}
    \includegraphics[trim={0.5cm, 0.5cm, 0.2cm, 0.5cm}, clip, width=0.24\columnwidth]{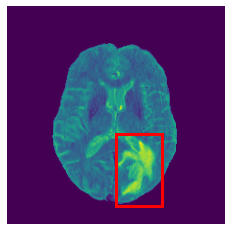}\hspace{0.01cm}
    \includegraphics[trim={0.5cm, 0.5cm, 0.2cm, 0.5cm}, clip, width=0.24\columnwidth]{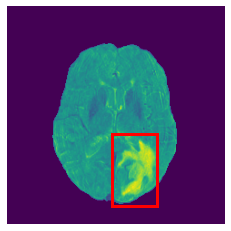}\hspace{0.01cm}
    \includegraphics[trim={0.5cm, 0.5cm, 0.2cm, 0.5cm}, clip, width=0.24\columnwidth]{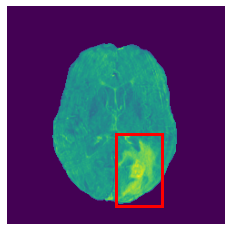}\hspace{0.01cm}

    \vspace{0.1cm}
    
%---- SAMPLE 2-----------
    {
    \includegraphics[trim={0.5cm, 0.5cm, 0.2cm, 0.5cm}, clip, width=0.24\columnwidth]{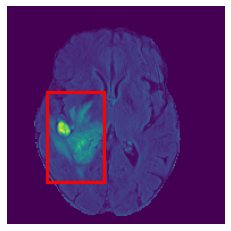}\hspace{0.01cm}
    \includegraphics[trim={0.5cm, 0.5cm, 0.2cm, 0.5cm}, clip, width=0.24\columnwidth]{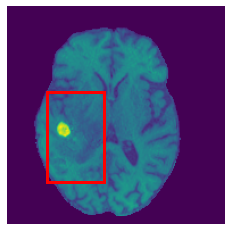}\hspace{0.01cm}
       \includegraphics[trim={0.5cm, 0.5cm, 0.2cm, 0.5cm}, clip, width=0.24\columnwidth]{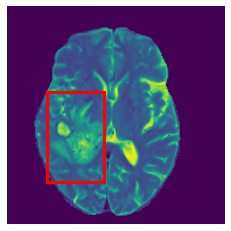}\hspace{0.01cm}
          \includegraphics[trim={0.5cm, 0.5cm, 0.2cm, 0.5cm}, clip, width=0.24\columnwidth]{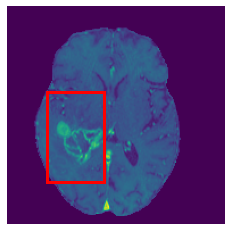}\hspace{0.01cm}
    \includegraphics[trim={0.5cm, 0.5cm, 0.2cm, 0.5cm}, clip, width=0.24\columnwidth]{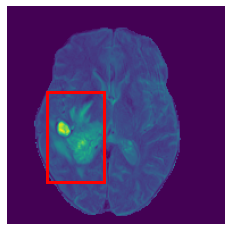}\hspace{0.01cm}
    \includegraphics[trim={0.5cm, 0.5cm, 0.2cm, 0.5cm}, clip, width=0.24\columnwidth]{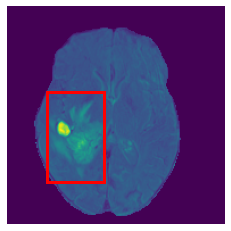}\hspace{0.01cm}
    \includegraphics[trim={0.5cm, 0.5cm, 0.2cm, 0.5cm}, clip, width=0.24\columnwidth]{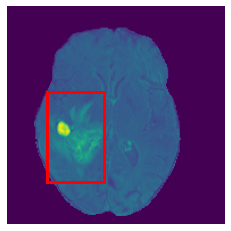}\hspace{0.01cm}
         
    \vspace{0.1cm}
    }
    
 %---- SAMPLE 3-----------
  {
    \includegraphics[trim={0.5cm, 0.5cm, 0.2cm, 0.5cm}, clip, width=0.24\columnwidth]{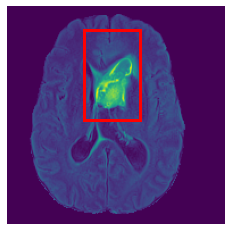}\hspace{0.01cm}
    \includegraphics[trim={0.5cm, 0.5cm, 0.2cm, 0.5cm}, clip, width=0.24\columnwidth]{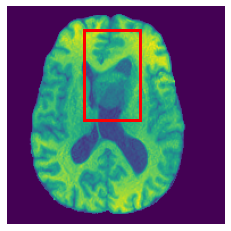}\hspace{0.01cm}
       \includegraphics[trim={0.5cm, 0.5cm, 0.2cm, 0.5cm}, clip, width=0.24\columnwidth]{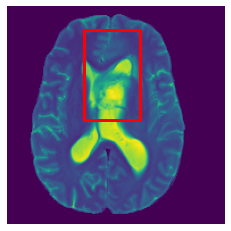}\hspace{0.01cm}
          \includegraphics[trim={0.5cm, 0.5cm, 0.2cm, 0.5cm}, clip, width=0.24\columnwidth]{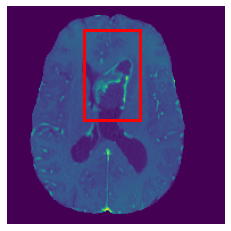}\hspace{0.01cm}
    \includegraphics[trim={0.5cm, 0.5cm, 0.2cm, 0.5cm}, clip, width=0.24\columnwidth]{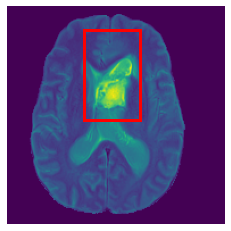}\hspace{0.01cm}
    \includegraphics[trim={0.5cm, 0.5cm, 0.2cm, 0.5cm}, clip, width=0.24\columnwidth]{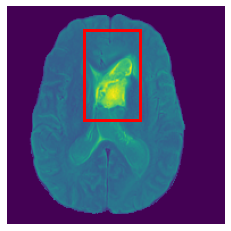}\hspace{0.01cm}
    \includegraphics[trim={0.5cm, 0.5cm, 0.2cm, 0.5cm}, clip, width=0.24\columnwidth]{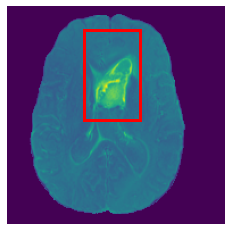}\hspace{0.01cm}
         
    \vspace{0.1cm}
    }
    
%---- SAMPLE 4-----------
    {
    \includegraphics[trim={0.5cm, 0.5cm, 0.2cm, 0.5cm}, clip, width=0.24\columnwidth]{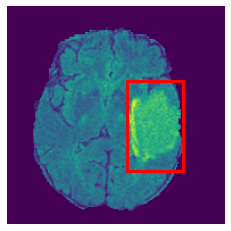}\hspace{0.01cm}
    \includegraphics[trim={0.5cm, 0.5cm, 0.2cm, 0.5cm}, clip, width=0.24\columnwidth]{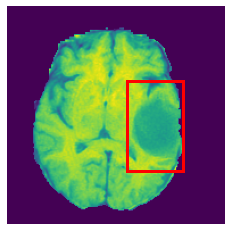}\hspace{0.01cm}
       \includegraphics[trim={0.5cm, 0.5cm, 0.2cm, 0.5cm}, clip, width=0.24\columnwidth]{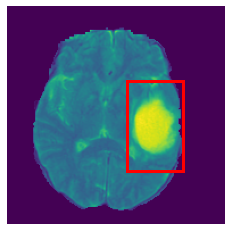}\hspace{0.01cm}
          \includegraphics[trim={0.5cm, 0.5cm, 0.2cm, 0.5cm}, clip, width=0.24\columnwidth]{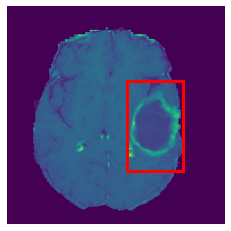}\hspace{0.01cm}
    \includegraphics[trim={0.5cm, 0.5cm, 0.2cm, 0.5cm}, clip, width=0.24\columnwidth]{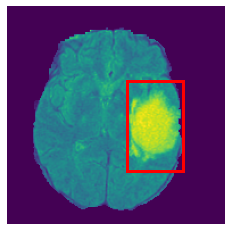}\hspace{0.01cm}
    \includegraphics[trim={0.5cm, 0.5cm, 0.2cm, 0.5cm}, clip, width=0.24\columnwidth]{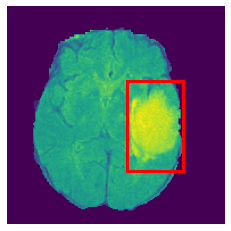}\hspace{0.01cm}
    \includegraphics[trim={0.5cm, 0.5cm, 0.2cm, 0.5cm}, clip, width=0.24\columnwidth]{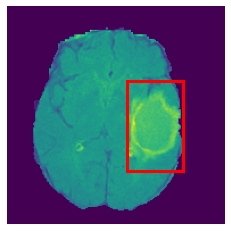}\hspace{0.01cm}
         
    \vspace{0.1cm}
    }
 \end{center}
    
\caption{{Tumor Enhancement: From $S_1$ to $S_4$ - uni-modality raw data in the order of FLAIR, T1, T2, and T1ce; $M_2$, $M_6$, and $M_9$ stand for the segmentation results obtained via multi-modality MRI embedding of $\{Flair, T2\}$,  $\{Flair, T1, T2\}$, and $\{Flair, T1, T2, T1ce\}$, respectively. Row 1 - 4 represent the HGG data collected from patient no. 1, 16, 56, and 112 from BraTS'19}. \textcolor{red}{$\rectangle$} - Whole tumor region.}
 \label{fig:sample_mri_embedding}
   
\end{figure*}

 The data preprocessing is a vital step to normalize intensities of brain MRI samples to have a similar distribution in order to avoid any initial bias. Especially for data-driven approaches, like supervised deep learning models, the data preprocessing is quite an important phase. Here, the following operations are carried out specifically on brain regions that are independent across modalities.
 
 \noindent\textcolor{gray}{$\triangle$} The MR images are skull stripped and co-registered to have a uniform resolution. 
 
 \noindent\textcolor{gray}{$\triangle$} The muti-channel MRI embedding is carried out using the four modalities Flair, T1, T2, and T1ce.
 
 \noindent\textcolor{gray}{$\triangle$} The top and bottom intensity percentiles are removed, and then data normalization is carried out by subtracting the mean and dividing it with the standard deviation values.

 \subsubsection{MRI Embedding}\label{mri_embedding}
 %----------------------------

The embedding processing is carried out using pixel-level fusion of the T1, T2, FLAIR, and T1ce modalities to enrich the information associated with each pixel of the brain MRI. 
The embedding process is governed by Eq.(~\ref{eq:embedding}) that requires the modalities to have the same depth, type and spatial resolution. 

\begin{equation}
    I_{embedded} = \frac{\alpha \cdot \mathbf{M_1} + \beta \cdot \mathbf{M_2} + \gamma \cdot \mathbf{M_3} + \lambda \cdot \mathbf{M_4}}{N} + c.
    \label{eq:embedding}
\end{equation}

In this case, $M_1, M_2, M_3$, and $M_4$ are the matrices that store the four MR modalities stated earlier, $\alpha, \beta, \gamma$, and $\lambda$ are the corresponding weights to the respective modalities to be considered wrt their individual credibility, and $N$ is the total number of modalities. The $c$ is a static weight that will be added to all the pixels of the embedded image to offset any hidden biases. 

For the simplicity and fast computation, this work sets $\alpha$, $\beta$, $\gamma$, and $\lambda$ to one and $c$ to zero. Note that if NumPy library is used the embedding will  be a pixel-wise simple addition with modulo operation, while in OpenCV library, it will be a simple addition with saturation. A sample of the multi-channel MRI-embedding outcome with three variation of channel mutations for four different patient data are shown in Fig.~\ref{fig:sample_mri_embedding} along the raw HGG input data. In which, one can observe that the tumor regions located by a \textcolor{red}{red} rectangles have been enhanced when the multi-channel MRI-embedding is applied. 

\subsubsection{Slice removal}\label{slice_removal}%**
Once the embedding process is completed, the first and the last few of the redundant slices are removed from the embedded input data, where each sample has a volume of 155. This work utilizes the slices from 30 to 120 and discards the rest of the blank and redundant slices. The new volume of slices are reshaped from $240 \times240$ to $192  \times 192$ in order to satisfy the input layer of U-Net. Note that the order of the operations embedding and slice removal plays a key role in optimizing required processing time. The impact of this order is analyzed in Section~\ref{preprocessing_analysis}.

\subsection{Phase 3: Model Training and Prediction}
%--------------------------------------

\subsubsection{The model}\label{themodel}%**
We have used the 9-layer U-Net architecture~\cite{polo8214}. The total  parameters of network is $31,054,145$, among them $31,042,369$ are trainable parameters and $11,776$ are non trainable parameters. All the convolutional operations are followed by a non-linear inclusion using Rectified Linear Unit (ReLU) function, except the last layer that employs a Sigmoid activation function to predict the whole tumor segmentation label. 

\subsubsection{Training}%**
The preprocessed data is split into mutually exclusive training and test with a ratio of $80:20$. For training time validation, a 20\% of random samples assigned into validation set from the training set itself. The model is trained from the scratch for $K$ number of epochs using Adam optimizer with a batch size of $b$, where $K$ and $b$ are set to $250$ and $32$, respectively. The objective function of the training process is given in Eq.~(\ref{eq:cost}).

\subsubsection{Prediction}%*
The trained model is evaluated on test set. The model's predictions from the top-layer with Sigmoid activator is compared against the ground truths of core tumors based on dice coefficient as defined by Eq.~(\ref{eq:dice}).

\begin{figure}[ht]
    \includegraphics[width=0.49\textwidth]{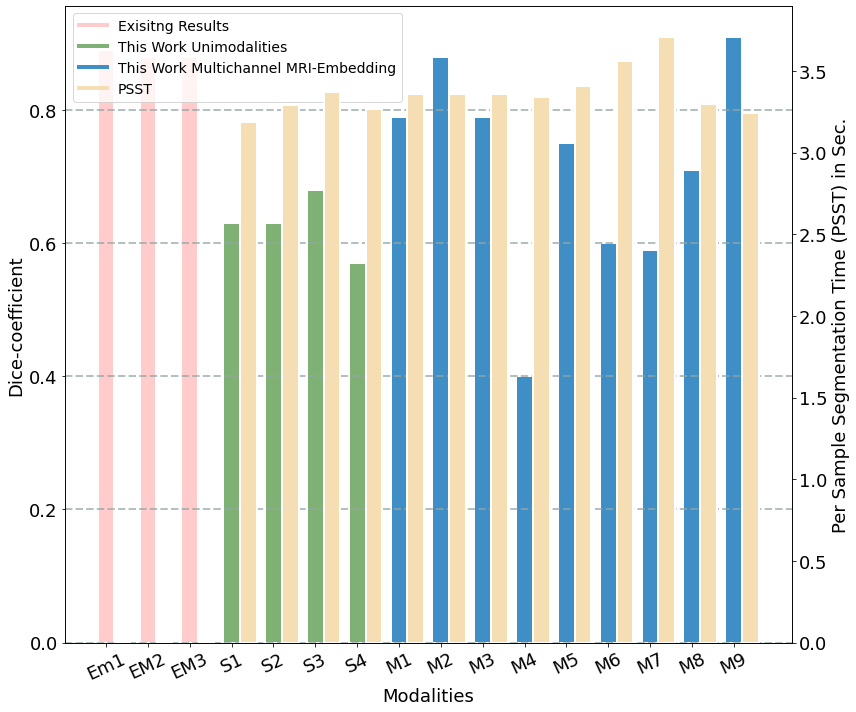}
    \caption{Performance Comparisons: EM1 to EM3 - existing models, S1 to S4 - unimodalities, and M1 to M9 - Multi-channel MRI-embedding modalities.}
    \label{fig:results_comparision_exsisting_vs_proposed}
\end{figure}

%-------------------------------
\section{Experimental Analysis}
%--------------------------------

\subsection{Environment}

The experimental study was carried out using the U-Net architecture as described in Section~\ref{themodel} written in Python 3 with Keras and Tesorflow libraries. The model was trained and tested on machine equipped with Intel hyper threaded 2.3 GHz Xeon processor and a Tesla K80 GPU having 2496 CUDA cores, and 35 GB DDR5 VRAM on Goole Colab.

\subsection{Dataset}

The experiments are conducted on the Brain Tumor Segmentation 2019 (BraTs'19) dataset provided by Medical Image Computing and Computer Assisted Intervention (MICCAI) society. 
It includes $285$ different cases among them, $210$ are HGG ($\mathbf{H} = \{h_1, h_2,\cdot\cdot\cdot, h_{210}\}$) and remaining $75$ are LGG ($\mathbf{L} = \{l_1, l_2,\cdot\cdot\cdot, l_{75}\}$).

Where, the samples in $\mathbf{H}$ and $\mathbf{L}$ are co-registered to the same anatomical template, interpolated to the same resolution of $1~mm^3$, and skull stripped. Hence, for every patient the dataset includes information of four channels, $\mathbf{C} =$ \{FLAIR, T1, T2, T1ce\}, where $c_{i} \in \Re^{240\times240\times155} $.

\subsection{Evaluation Metrics}
%------------------------------
This study utilizes the dice coefficient defined in Eq.~\eqref{eq:dice} as the evaluation metric that penalizes the false positive and the false negative values.
 
It is the most commonly used metric in biomedical image segmentation and its measure states the similarity between predicted and human annotated segmentations. 

\begin{equation}
Dice = \frac{2 \cdot TP}{2 \cdot TP+FN+FP},
\label{eq:dice}
\end{equation}

\begin{equation}
Dice~Loss = 1 - \frac{2 \cdot TP}{2 \cdot TP+FN+FP},
\label{eq:cost}
\end{equation}
where $TP, FN,$ and $FP$ stand for True positive, False negative, and False positive, respectively.

\subsection{Step-by-Step Performance Analysis}

%---------------------------------

This Section provides a systematic evaluation of the proposed approach. 

\subsubsection{Impact of the Preprocessing Order}\label{preprocessing_analysis}

\begin{table}[t]
\caption{Impact of preprocessing order: Average Per Sample Time taken for MRI Embedding (\ref{mri_embedding}) Before and After Slice Removal (\ref{slice_removal}) in Millisecond.}
\begin{center}
\begin{tabular}{|p{3.5cm}|c|c|}
\hline\hline

\multirow{2}{3.5cm}{\centering \textbf{Multi-channel MRI-embedding Models}} & \multicolumn{2}{|c|}{\textbf{Preprocessing Order}} \\ \cline{2-3}
& \textbf{\ref{mri_embedding} $\rightarrow$ \ref{slice_removal}} & \textbf{\ref{slice_removal} $\rightarrow$ \ref{mri_embedding}} \\ 

 \hline\hline

$M_1$~-~Flair, T1 & {6412.44} & {6385.71} \\\hline
$M_2$~-~Flair, T2 & {6416.42} & {6451.95} \\\hline
$M_3$~-~Flair, T1CE & {6483.36} & {6503.98} \\\hline
$M_4$~-~T1, T1CE & {6380.00}  & {6480.70} \\\hline
$M_5$~-~T2, T1CE & {6447.17}  & {6464.07} \\\hline
$M_6$~-~Flair, T1, T2 & {9632.06}  & {9648.12} \\\hline
$M_7$~-~Flair, T1, T1CE & {9710.22}  & {9730.62} \\\hline
$M_8$~-~Flair, T2, T1CE & {9709.33}  & {9756.86} \\\hline
$M_9$~-~Flair, T1, T2, T1CE & {12868.46} & {13071.13} \\ \hline

\textcolor{blue}{Overall} & \textcolor{blue}{74059.46} & \textcolor{blue}{74492.84} \\\hline\hline

\end{tabular}
\label{tab:preprocessing_time}
\end{center}
\end{table}

\begin{table}[t]
\begin{center}
\caption{Performance of various models on BraTs'19 and their \% of improvement in dice coefficient wrt the best models \cite{ziqi}: ${\downarrow}$ and ${\uparrow}$ stand for negative and positive improvements, respectively}
\begin{tabular}{|l|c|c|}
\hline\hline
\textbf{\centering{Modalities/Model}} & {\textbf{{Dice Coef}} } & \textbf{\% Improvement}\\\hline\hline

\multicolumn{3}{|c|}{\textbf {Existing Models} }\\\hline

$EM_1$ - DAN w/ CRF~\cite{ziqi} & \underline{0.89} & - ${\mathrm{}}$ \\\hline
$EM_2$ - 3D U-Net~\cite{li2019multi} & 0.88 & ${\downarrow}$ 1~\% ${\mathrm{}}$ \\ \hline
$EM_3$ - Cascaded U-Net~\cite{lachinov2020knowledge} & 0.88 & ${\downarrow}$ 1~\% ${\mathrm{}}$\\ \hline

\multicolumn{3}{|c|}{\textbf{This work: Uni-modality w/ U-Net}} \\ \hline

$S_1$ - Flair  & 0.63 & ${\downarrow}$ 26~\% ${\mathrm{}}$  \\ \hline
$S_2$ - T1 & 0.63 & ${\downarrow}$ 26~\% ${\mathrm{}}$ \\ \hline
$S_3$ - T2 & \underline{0.68} & ${\downarrow}$ 21~\% ${\mathrm{}}$ \\ \hline
$S_4$ - T1ce & 0.57 & ${\downarrow}$ 32~\% ${\mathrm{}}$ \\ \hline

\multicolumn{3}{|c|}{\textbf{This work: Multi-channel MRI-embedding w/ U-Net}} \\ \hline

$M_1$ - Flair, T1 & 0.79 & ${\downarrow}$ 10~\% ${\mathrm{}}$  \\\hline
$M_2$ - Flair, T2 & 0.88 & ${\downarrow}$ 1~\% ${\mathrm{}}$  \\\hline
$M_3$ - Flair, T1ce & 0.79 & ${\downarrow}$ 10~\% ${\mathrm{}}$  \\\hline
$M_4$ - T1, T1ce & 0.40 & ${\downarrow}$ 48~\% ${\mathrm{}}$  \\\hline
$M_5$ - T2, T1ce & 0.75 & ${\downarrow}$ 14~\% ${\mathrm{}}$  \\\hline
$M_6$ - Flair, T1, T2 & 0.60 & ${\downarrow}$ 29~\% ${\mathrm{}}$  \\\hline
$M_7$ - Flair, T1, T1ce & 0.59 & ${\downarrow}$ 30~\% ${\mathrm{}}$  \\\hline
$M_8$ - Flair, T2, T1ce & 0.71 & ${\downarrow}$ 18~\% ${\mathrm{}}$  \\\hline
\textcolor{blue}{$\mathbf{M_9}$ - \textbf{Flair, T1, T2, T1ce}} & \textcolor{blue}{\textbf{0.91}} & \textcolor{blue}{\textbf{${\uparrow}$}\textbf{ 2~\%}} ${\mathrm{}}$  \\
\hline\hline
\end{tabular}
\label{tab:performance_comparision}
% \vspace{0.5cm}
\end{center}
\end{table}

\begin{table*}[t]
    \centering
    \caption{Average Per Sample Segmentation Time (PSST) of Various Modalities in $s$.}
    \label{tab:timing_analysis_single_vs_multimodality_embedding}
    \begin{tabular}{|c|c|c|c|c|c|c|c|c|c|c|c|c|c|c|}
    \hline
       Modalities  &  $S_1$  &  $S_2$    &  $S_3$  &  $S_4$  &  $M_1$  &  $M_2$  &  $M_3$  & $M_4$  &  $M_5$  & $M_6$  & $M_7$  &  $M_8$  &  $M_9$ \\
       \hline\hline
        PSST & 3.168 & 3.289 &  3.368  &  3.270  & 3.358  &  3.356  & 3.361  &  3.342  &  3.412  & 3.561  &  3.371  &  3.301  &  3.243\\
         \hline\hline
    \end{tabular}
    
\end{table*}

\begin{figure*}[ht]
  
    \begin{tabular}{p{1.8cm}p{1.8cm}p{2cm}p{1.5cm}p{2.1cm}p{1.8cm}p{1.6cm}p{2.0cm}}
       \centering GT  &  \centering $S_1$ & \centering $S_2$ & \centering $S_3$ & \centering $S_4$ & \centering $M_2$ & \centering $M_6$ & \centering $M_9$\\
    \end{tabular}
    \vspace{-0.5cm}
\begin{center}
%---- SAMPLE 1-----------
    {\includegraphics[trim={0.5cm, 0cm, 0.2cm, 0.9cm}, clip,width=0.24\columnwidth]{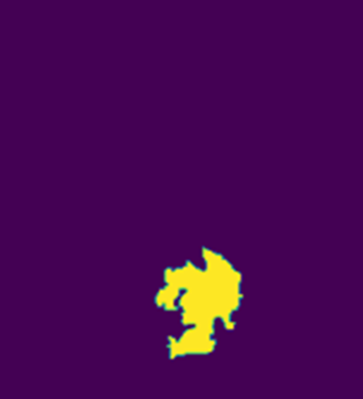}\hspace{0.01cm}
    \includegraphics[trim={0.5cm, 0cm, 0.2cm, 0.9cm}, clip, width=0.24\columnwidth]{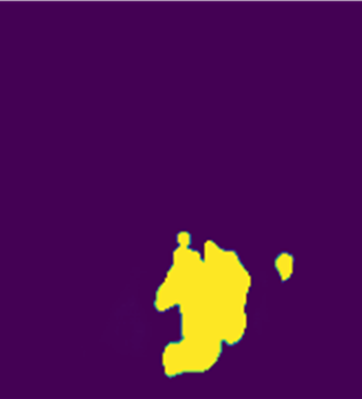}\hspace{0.01cm}
    \includegraphics[trim={0.2cm, 0.15cm, 0.4cm, 0.9cm}, clip,width=0.24\columnwidth]{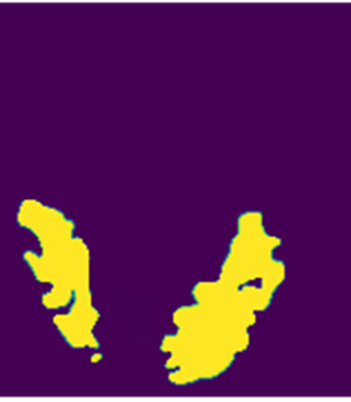}\hspace{0.01cm}
    \includegraphics[trim={0.5cm, 0cm, 0.2cm, 0.9cm}, clip,  width=0.24\columnwidth]{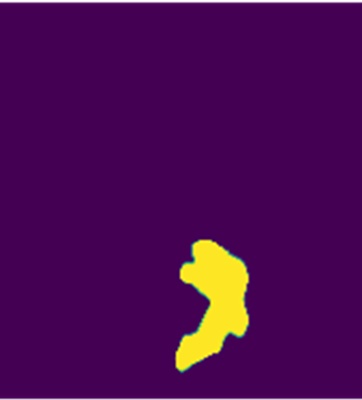}\hspace{0.01cm}
    {\includegraphics[trim={0.5cm, 0.1cm, 0.2cm, 0.6cm}, clip,  width=0.24\columnwidth]{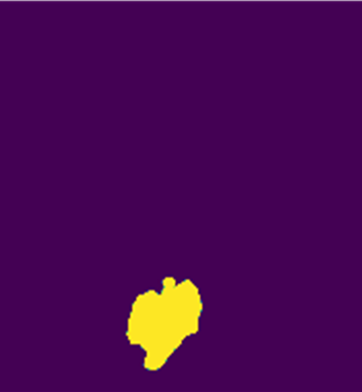}}\hspace{0.01cm}
    \includegraphics[trim={0.5cm, 0cm, 0.2cm, 0.8cm}, clip, width=0.24\columnwidth]{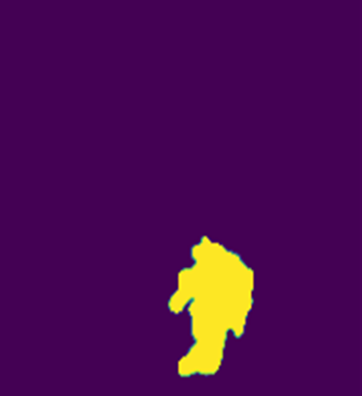}\hspace{0.01cm}
    \includegraphics[trim={0.9cm, 0.3cm, 0.2cm, 0.1cm}, clip, width=0.24\columnwidth]{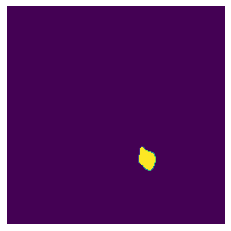}\hspace{0.01cm}
    \includegraphics[trim={0.5cm, 0cm, 0.2cm, 0.85cm}, clip, width=0.24\columnwidth]{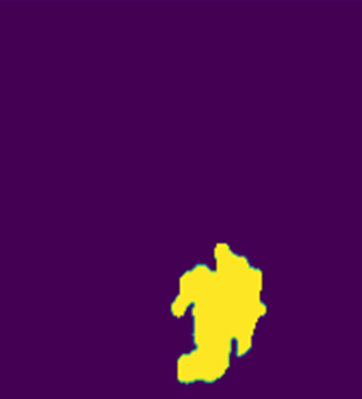}\hspace{0.01cm}
    }
    \vspace{0.1cm}
    
%---- SAMPLE 2-----------
    {\includegraphics[trim={0.5cm, 0cm, 0.2cm, 0.3cm}, clip, width=0.24\columnwidth]{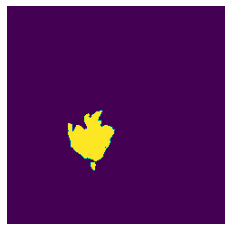}\hspace{0.01cm}
        \includegraphics[trim={0.5cm, 0cm, 0.2cm, 0.3cm}, clip, width=0.24\columnwidth]{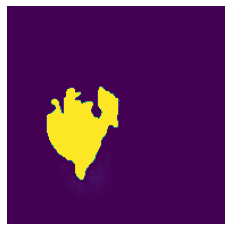}\hspace{0.01cm}
            \includegraphics[trim={0.5cm, 0cm, 0.2cm, 0.3cm}, clip, width=0.24\columnwidth]{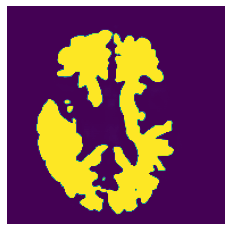}\hspace{0.01cm}
                \includegraphics[trim={0.5cm, 0cm, 0.2cm, 0.3cm}, clip, width=0.24\columnwidth]{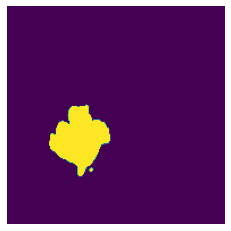}\hspace{0.01cm}
                    \includegraphics[trim={0.5cm, 0cm, 0.2cm, 0.3cm}, clip, width=0.24\columnwidth]{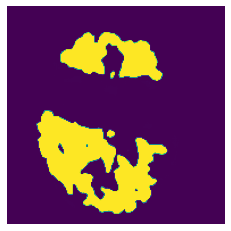}\hspace{0.01cm}
            \includegraphics[trim={0.5cm, 0cm, 0.2cm, 0.3cm}, clip, width=0.24\columnwidth]{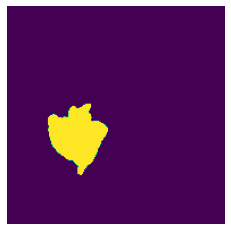}\hspace{0.01cm}
                \includegraphics[trim={0.5cm, 0cm, 0.2cm, 0.3cm}, clip, width=0.24\columnwidth]{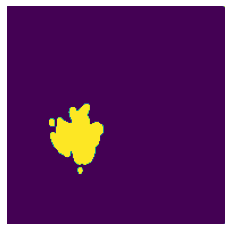}\hspace{0.01cm}
            \includegraphics[trim={0.5cm, 0cm, 0.2cm, 0.3cm}, clip, width=0.24\columnwidth]{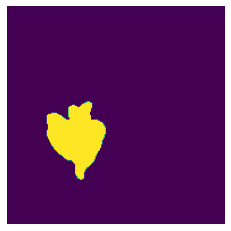}\hspace{0.01cm}
    }
    \vspace{0.1cm}
    
 %---- SAMPLE 3-----------
    {\includegraphics[trim={0.5cm, 0cm, 0.2cm, 0.3cm}, clip, width=0.24\columnwidth]{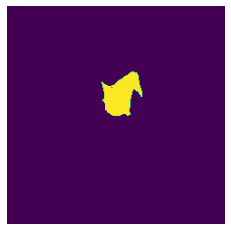}\hspace{0.01cm}
        \includegraphics[trim={0.5cm, 0cm, 0.2cm, 0.3cm}, clip, width=0.24\columnwidth]{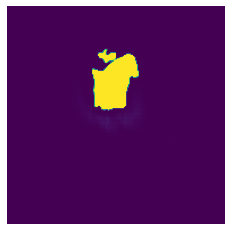}\hspace{0.01cm}
            \includegraphics[trim={0.5cm, 0cm, 0.2cm, 0.3cm}, clip, width=0.24\columnwidth]{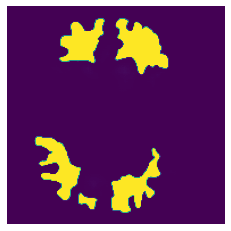}\hspace{0.01cm}
                \includegraphics[trim={0.5cm, 0cm, 0.2cm, 0.3cm}, clip, width=0.24\columnwidth]{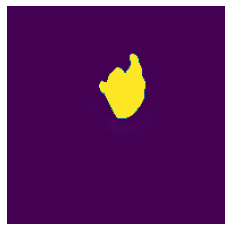}\hspace{0.01cm}
                    \includegraphics[trim={0.5cm, 0cm, 0.2cm, 0.3cm}, clip, width=0.24\columnwidth]{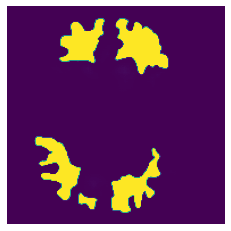}\hspace{0.01cm}
            \includegraphics[trim={0.5cm, 0cm, 0.2cm, 0.3cm}, clip, width=0.24\columnwidth]{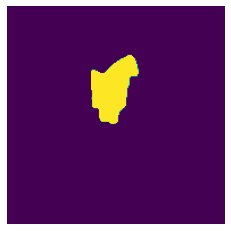}\hspace{0.01cm}
                \includegraphics[trim={0.5cm, 0cm, 0.2cm, 0.3cm}, clip, width=0.24\columnwidth]{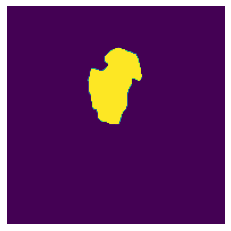}\hspace{0.01cm}
            \includegraphics[trim={0.5cm, 0cm, 0.2cm, 0.3cm}, clip, width=0.24\columnwidth]{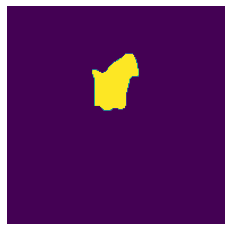}\hspace{0.01cm}
    } 
    \vspace{0.05cm}
    
%---- SAMPLE 4-----------
    {\includegraphics[trim={0.5cm, 0cm, 0.2cm, 0.3cm}, clip, width=0.24\columnwidth]{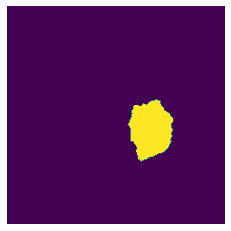}\hspace{0.01cm}
        \includegraphics[trim={0.5cm, 0cm, 0.2cm, 0.3cm}, clip, width=0.24\columnwidth]{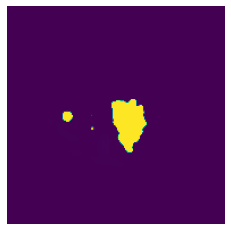}\hspace{0.01cm}
            \includegraphics[trim={0.5cm, 0cm, 0.2cm, 0.3cm}, clip, width=0.24\columnwidth]{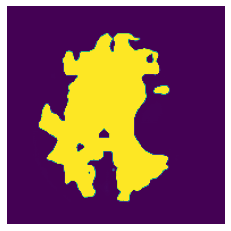}\hspace{0.01cm}
                \includegraphics[trim={0.5cm, 0cm, 0.2cm, 0.3cm}, clip, width=0.24\columnwidth]{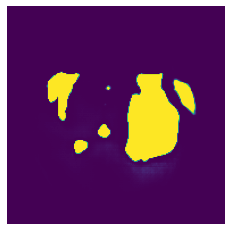}\hspace{0.01cm}
                    \includegraphics[trim={0.5cm, 0cm, 0.2cm, 0.3cm}, clip, width=0.24\columnwidth]{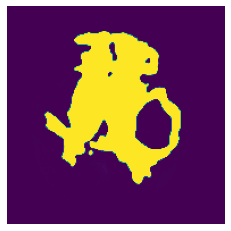}\hspace{0.01cm}
            \includegraphics[trim={0.5cm, 0cm, 0.2cm, 0.3cm}, clip, width=0.24\columnwidth]{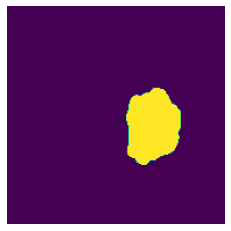}\hspace{0.01cm}
                \includegraphics[trim={0.5cm, 0cm, 0.2cm, 0.3cm}, clip, width=0.24\columnwidth]{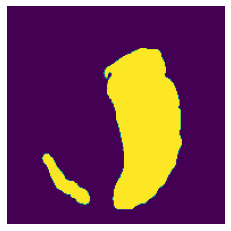}\hspace{0.01cm}
            \includegraphics[trim={0.5cm, 0cm, 0.2cm, 0.3cm}, clip, width=0.24\columnwidth]{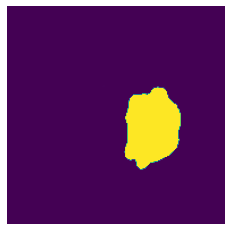}\hspace{0.01cm}
    } 
    \vspace{0.1cm}
 \end{center}
 \vspace{-0.5cm}   
\caption{{Sample of Qualitative Results: $GT$ - Ground truth; $S_1$ - $S_4$ Stand for the Results Obtained using the Uni-modality inputs, Flair, T1, T2 and T1ce respectively; $M_1$ - $M_9$ Represent the Results Obtained via Multi-channel MRI-embedding Inputs, \{Flair, T2\},  \{Flair, T1, T2\}, and \{Flair, T1, T2, T1ce\}, respectively. {Row 1 - 4 Are the Results for the HGG MRI's of the Patient No. 1, 16, 56, and 112 in the BraTS'19}.}}
 \label{fig:qualitative_results}
   
\end{figure*}

Table~\ref{tab:preprocessing_time} summaries the timing complexity of the preprocessing order of the proposed multi-channel MRI-embedding and slice removal across different set of modalities as an average time for one patient 
The analysis shows that order of the preprocessing plays an important role in minimizing the required time such that one should carry out the slice removal-\ref{slice_removal} after the MRI embedding-\ref{mri_embedding} for a quicker preprocessing. When the order is reversed, the slice removal needs to be done on each modality individually to maintain a equal volume in all the channels, then only the MRI embedding can be completed. However, if the MRI embedding is performed at first, there should not be any problem of maintaining the equal volume in each channel since original data has the same depth of 155 on all modalities. Thus, in this way the slice removal has to be done only once on the embedded (fused) input to the Phase 2 shown in Fig.~\ref{Fig:flow_diagram}. Thus, carrying out the preprocessing in the right order of MRI embedding $\rightarrow$ slice removal in stead of slice removal $\rightarrow$ MRI embedding, there is a possibility of saving 1\% of per sample processing time.

\subsubsection{Quantitative Analysis}\label{seg_analysis}
%----------------------------------------------
Quantitative analysis is performed in two parts that are single modality using individual raw data inputs and proposed multi-modality MRI-embedding-based inputs and the results are compared in Table~\ref{tab:performance_comparision} and in Fig.~\ref{fig:results_comparision_exsisting_vs_proposed} with the best literature in recent times. Among the existing approaches, the DAN w/ CRF~\cite{ziqi} model achieves the best results of 0.89 dice-coefficient. Then, taking this as a baseline, rest of the models' performances are compared. From the ablation study, it can be derived that using a uni-channel or unimodality input data causes a poor segmentation outcome; among them, the T2 modality has been the best. For instance, the individual channels T1 and T1-ce produce 0.63 and 0.57 dice-coefficient, respectively. The characteristic of these two modalities is to determine inner part of whole tumor excluding the edema; thus, it results poor segmentation of the whole tumor. However, the other two channels, namely, T2 and Flair, contain the crucial information of the outer boundary of the whole tumor. 
Based on that, the various levels of all the channels mutation process through the proposed MRI-embedding whenever the T2 and Flair data are involved. For example, for the MRI-embedding of Flair with T1 and T2 with T1ce, the whole tumor segmentation result is improved by $16\%$ and $12\%$ when compared to the best of uni-channel, i.e., T2-based segmentation. However, these performances are still lower than the literature by $10\%$ and $14\%$, respectively. This condition is improved when the proposed multi-channel MRI-embedding is deployed with full-level mutation of all the channels ($\mathbf{M_9}$ - {Flair, T1, T2, T1ce} in Table~\ref{tab:performance_comparision}) and gains $2\%$ higher dice-coefficient than the best existing model, DAN w/ CRF~\cite{ziqi}.  

In terms of per sample prediction time (refer to Table~\ref{tab:timing_analysis_single_vs_multimodality_embedding} and Fig.~\ref{fig:results_comparision_exsisting_vs_proposed}), there the difference between different modalities is negligible since number of slices used in all the modalities do not change. However, the slight variation can be caused by the pixel-level and global-level intensity changes in the input modalities due to the difference in MR imaging technology and level of mutation in the MRI-embedding.

\subsection{Qualitative Analysis}
%--------------------------------
A qualitative analysis is carried through visual inspection of the segmented results in comparison to the respective ground truths. Considering the page limitation of the paper, some qualitative results of four patients using has uni-channels and three multi-channel MRI-embedding are shown in Fig.~\ref{fig:qualitative_results} along with their human annotated ground truths of the patients' whole tumor. From observation, it can be concluded that the proposed MRI-embedding models, $M_2$ and $M_9$ has identified the whole tumor close to the expert annotations. 

\section{Conclusion}
%---------------------

This work presents a simple yet effective strategy called multi-channel MRI-embedding for improving the whole tumor segmentation. The experimental analysis on the benchmark BraTs-2019 dataset shows that proposed model achieves the best performance when compared to the existing approaches. The future direction is dedicated for exploring other means of MRI-embedding techniques, for instance, using subspace transformations and weighted grading.

\section*{Acknowledgment}

This work acknowledges the Google for generosity of providing the HPC on the Colab machine learning platform.

\bibliographystyle{./bibliography/IEEEtran}
\bibliography{./bibliography/IEEEabrv,./bibliography/IEEEexample}

\vspace{12pt}

\end{document}